\documentclass[conference]{IEEEtran}
\IEEEoverridecommandlockouts
\usepackage{cite}
\usepackage{amsmath,amssymb,amsfonts}
\usepackage{algorithmic}
\usepackage{graphicx}
\usepackage{textcomp}
\usepackage{xcolor}
\usepackage[compatibility=false]{caption}
\usepackage{subcaption}
\usepackage{booktabs} 

\usepackage[most]{tcolorbox}
\usepackage{url,float}
\definecolor{navyref}{RGB}{0,32,96}
\usepackage[colorlinks=true, linkcolor=navyref, citecolor=navyref, urlcolor=navyref]{hyperref}
\hypersetup{
    pdftitle={AlphaRoute: Large Language Models as Semantic Optimizers for Multi-Objective Routing},
    pdfauthor={Kabir Murjani, Mishri Bhavsar, Manish I. Patel, Jonti Talukdar},
    pdfsubject={Combinatorial Optimization},
    pdfkeywords={Combinatorial Optimization, Global Routing, Physical Design, Rip-up and Reroute, Large Language Models, Knowledge Graphs, Explainable AI}
}

\def\BibTeX{{\rm B\kern-.05em{\sc i\kern-.025em b}\kern-.08em
    T\kern-.1667em\lower.7ex\hbox{E}\kern-.125emX}}

\begin{document}

% IEEE catalog number (979-8-3195-1246-8/26/$31.00 (c)2026 IEEE) intentionally
% omitted: it identifies the IEEE-published version of record, which IEEE's
% posting policy does not permit on arXiv. Only the accepted version may be
% posted. Restore it if building the camera-ready for the proceedings.

\title{AlphaRoute: Large Language Models as Semantic Optimizers for Multi-Objective Routing
}

\author{
    Kabir Murjani\textsuperscript{1, *}\thanks{* Corresponding author: \href{mailto:kabirmurjani@gmail.com}{kabirmurjani@gmail.com}}\thanks{The complete codebase and evaluation scripts required to reproduce all results are publicly available at \protect\url{https://github.com/Kcbir/AlphaRoute}. The benchmarks are the public ISPD 2025 contest designs \cite{ispd2025contest}.}\thanks{\copyright~2026 IEEE. Personal use of this material is permitted. Permission from IEEE must be obtained for all other uses, in any current or future media, including reprinting/republishing this material for advertising or promotional purposes, creating new collective works, for resale or redistribution to servers or lists, or reuse of any copyrighted component of this work in other works.}, Mishri Bhavsar\textsuperscript{2}, Manish I. Patel\textsuperscript{2}, Jonti Talukdar\textsuperscript{3} \\

    %Kabir Murjani\textsuperscript{1}, Mishri Bhavsar\textsuperscript{2}, Manish I. Patel\textsuperscript{2}, Jonti Talukdar\textsuperscript{3} \\

    \small
    \textsuperscript{1}Department of Electrical Engineering, Institute of Technology, Nirma University, Ahmedabad, India \\
    \textsuperscript{2}Department of Electronics and Communication Engineering, Institute of Technology, Nirma University, Ahmedabad, India \\
    %\textsuperscript{3}Department of Electrical and Computer Engineering, Duke University, Durham, NC, USA
    %\textsuperscript{3}ASU Center for Semiconductor Microelectronics, Tempe, AZ
    \textsuperscript{3}Center for Semiconductor Microelectronics, Arizona State University, Tempe, Arizona, USA
}
\maketitle
\begin{abstract}
Very Large Scale Integration (VLSI) global routing is an NP-hard combinatorial optimization problem requiring signal net assignment across capacity-constrained 3D grids while minimizing congestion, wirelength, and via transitions. Because traditional heuristics rely on static penalty schedules that fail on complex congestion topologies, we present AlphaRoute: a multi-objective adaptive search framework reformulating rip-up and reroute (R\&R) into a dynamic optimization system. We introduce SHAP-based overflow decomposition to isolate per-net congestion, driving targeted subgraph extraction via 3D Dijkstra maze routing and an adaptive PathFinder policy. Crucially, AlphaRoute employs Large Language Models (LLMs) as semantic policy optimizers. Bounded by a deterministic knowledge graph, the LLMs interpret congestion metrics to dynamically adjust penalty parameters. Evaluated on ISPD 2025 benchmarks, AlphaRoute reduces overflow by 98.6\% on MEMPOOL. On the constrained ARIANE design, we achieve an overflow of 146,109 (a 29.8x reduction in overflow over the state of the art), yielding a penalized score of $S_{orig}=0.0538$ versus the State-of-the-art (SOTA) 1.780. These results demonstrate that superior algorithmic search geometry can overcome the latency of interpreted Python implementations.
\end{abstract}


\section{Introduction}

Global routing is an important phase of Very Large Scale Integration (VLSI) physical design that assigns coarse routing paths for electrical connections (nets) across a chip layout, followed by detailed routing, which determines the exact routing tracks for each net. The chip is partitioned into global routing cells (GCells), and each signal net is assigned to a sequence of GCells to minimize wire length, reduce congestion, and control via count. Poorly managed congestion at this stage propagates into downstream violations: unroutable designs with Design Rule Checking (DRC) failures, timing degradation, excess power draw, and inflated die area. Modern designs intensify this challenge. Contemporary System-on-Chips (SoCs) contain hundreds of thousands to millions of nets routed across ten or more metal layers within grids exceeding one million GCells. At this scale, congestion interactions are highly non-local: a single over-subscribed channel can force cascading detours that ripple across distant regions of the chip. Classical routers manage these interactions through fixed penalty schedules, yet the optimal penalty trajectory depends on the specific congestion topology encountered at each iteration, information that static heuristics cannot exploit. Furthermore, as routing frameworks grow in complexity, designers lose visibility into which algorithmic knob drove a particular outcome, making it difficult to diagnose failures or transfer insights across designs.

The global routing problem is NP-hard \cite{garey1977rectilinear}, so practical solvers rely on heuristics. CUGR \cite{liu2020cugr} targets detailed routability through probabilistic resource modeling. InstantGR \cite{lin2025instantgr} reduces runtime through GPU parallelization. PathFinder-based routers \cite{mcmurchie1995pathfinder} use negotiation-based congestion resolution to balance overflow elimination against critical-path delay, and recent accelerators such as PAGR \cite{solovyev2025pagr} target throughput of the global routing flow itself. Despite this progress, three limitations persist across existing frameworks: penalty parameters are static and not adapted to the evolving congestion landscape, rip-up candidate selection is not driven by per-net overflow contribution, and there is no interpretable reasoning layer that explains why a particular routing policy was chosen. This work addresses all three gaps. The specific research contributions are:

\begin{enumerate}
    \item SHapley Additive exPlanations (SHAP)-inspired overflow decomposition and targeted rip-up: We introduce a permutation-importance-style analyzer \cite{lundberg2017unified} that decomposes post-routing overflow into per-net scores, layer-wise fractions, and utilization-overflow correlations, enabling the rip-up stage to select exactly the nets responsible for congestion rather than relying on coarse heuristics.
    \item Adaptive PathFinder policy with capacity-aware routing: We extend the classical PathFinder cost model \cite{mcmurchie1995pathfinder} with SHAP-driven parameter rules that adjust the history penalty, present overflow weight, and via cost at each iteration. Initial routing further benefits from flexible dual L-shape evaluation and dynamic layer selection, which together reduce initial overflow by 2.2x on the ARIANE benchmark.
    \item Dijkstra maze rerouting within a 3D bounding-box subgraph: Medium-area nets ripped by the SHAP analyzer are rerouted through a cost-aware Dijkstra search over a local subgraph, allowing the router to detour around congestion hotspots that simple L-routes or Steiner trees cannot resolve.
    \item LLM-based interpretable policy adaptation with knowledge-graph validation: A Large Language Model (LLM) receives structured SHAP metrics and proposes parameter adjustments with natural-language reasoning. Proposals are validated against hard constraints stored in a RustworkX knowledge graph before application, combining interpretability with safety.
    
\end{enumerate}

On the International Symposium on Physical Design (ISPD) 2025 ARIANE benchmark dataset, AlphaRoute achieves an overflow of 146,109 (29.8x below SOTA in overflow) and a score of $S_{orig}=0.0538$ versus SOTA 1.780 (a $33.1\times$ score reduction).

\begin{figure}[t]
    \centering
     \includegraphics[height=0.62\textheight]{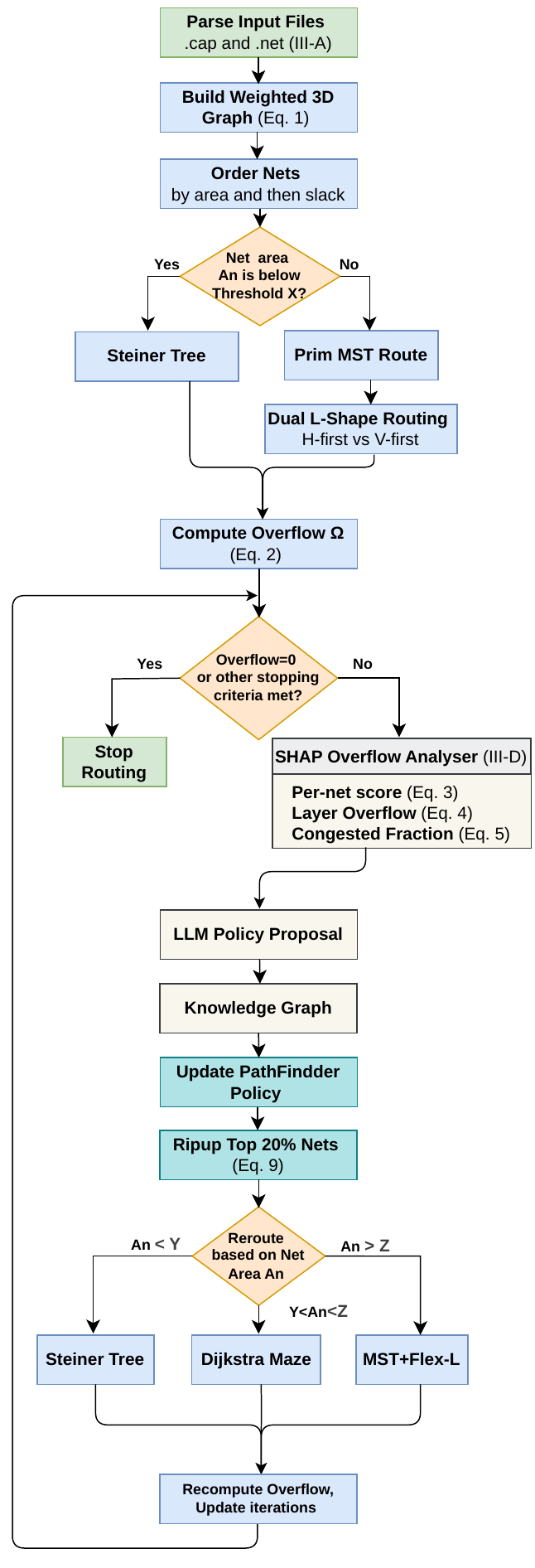}
    \caption{AlphaRoute execution flow: input parsing and graph construction, slack-aware net ordering, initial routing, and the SHAP-guided rip-up/reroute loop with knowledge-graph validated LLM policy adaptation.}
    \label{fig:Flow}
\end{figure}

\section{Related Work}

\subsection{Classical and Graph-Based Global Routing}
FLUTE \cite{chu2008flute} remains a standard for fast rectilinear Steiner tree construction, but Steiner methods reason locally per net, ignoring the multi-net resource contention that drives overflow. Graph-based learning methods address this gap: graph attention networks \cite{velickovic2018graph} introduced learned neighborhood weighting via message passing, and graph-representation Reinforcement Learning (RL) was shown to scale to industrial placement without manual feature engineering \cite{mirhoseini2021graph}. These results indicate that graph encoders can capture congestion coupling and long-range dependency more naturally than hand-crafted features.

\subsection{LLM-Based Evaluation and Its Limitations}
Using LLMs as automated judges scales evaluation but introduces systematic bias. A study of LLM judges found persistent position bias, verbosity bias, and self-preference, with the largest disagreements against human annotators on open-ended responses \cite{zheng2023judging}. In a routing loop, such bias becomes a training target that pushes policy toward style over substance. This motivates our use of LLMs strictly for structured policy suggestion rather than as reward judges.

\subsection{Reinforcement Learning from Human Feedback (RLHF) and Reward Grounding}
RLHF fits a reward model on human preferences and optimizes the policy against that signal \cite{ouyang2022training}, but the reward model must extrapolate beyond its training comparisons, and this error grows as the policy diverges from the data distribution \cite{gao2023scaling}. Nearest-neighbor language modeling showed that retrieval can anchor predictions to observed data instead of forcing unsupported generalization \cite{khandelwal2020generalization}. Our knowledge graph applies the same principle: it ties routing decisions to previously observed outcomes, keeping the loop grounded in measured results.

\section{Methodology}

This section describes AlphaRoute in order of execution, following the architecture in Fig.~\ref{fig:Flow}. The router, input parser, and RL policy loop are implemented in Python, with routing quality scored by a C++ evaluator.

\subsection{Input Representation and Graph Construction}
Input files (\texttt{.cap}, \texttt{.net}) are sourced from the ISPD 2025 contest
dataset \cite{ispd2025contest} and parsed. The \texttt{.cap} file
defines per-GCell routing capacities $\mathbf{C} \in \mathbb{R}^{N_L \times y_{\max}
\times x_{\max}}$, layer costs, and scoring weights over a $10 \times 646 \times 646$
grid (ARIANE); metal layer~0 is via-only. The \texttt{.net} file lists pin locations,
GCell access points, and pre-routing slack estimates. 
A RustworkX graph is constructed over the parsed grid.
Each GCell is mapped to a node via $\text{idx}(z,y,x) = z \cdot Y_{\max} \cdot X_{\max} + y \cdot X_{\max} + x$.
Edges encode legal routing moves: horizontal layers connect $(x,y,z)$ to
$(x{\pm}1,y,z)$, vertical layers connect it to $(x,y{\pm}1,z)$, and vias to $(x,y,z{\pm}1)$
for $z>1$ with a fixed cost of 40. The initial weight assigned to an edge $e$ between adjacent cells $u$ and $v$ on layer $z$ is:

\begin{equation}
W(e)=(10.0-3.0\cdot\sigma_{n})+(11-c_{z,u})+(11-c_{z,v})
\label{eq:base_weight}
\end{equation}

where $\sigma_{n}\in[0,1]$ is the criticality factor of the net, derived from the worst-case pin slack $s_{n}$ relative to the global worst slack $s_{min}$ as $\sigma_{n}=\max(0,1-s_{n}/s_{min})$. The net having slack as $s_{min}$ is the most time-critical net with a factor of $\sigma_{n}=1.0$ and a final base cost of 7.0, giving it a 30\% discount over a non-critical net. The terms $(11-c_{z,u})$ and $(11-c_{z,v})$ add between 1 (empty cell, $c=10$) and 11 (fully congested, $c=0$), routing traffic away from overloaded GCells. 

\subsection{Slack Aware Net Ordering}
Nets are ordered by bounding-box area first (large nets route first to reduce future congestion), then by most-negative slack among similarly sized nets, and finally single-GCell nets requiring only trivial resources. Area thus takes precedence over criticality: a large non-critical net routes before a smaller critical one, and ties in area are broken by criticality so the most time-critical net claims the shortest, least-resistive path. Non-critical nets route later and naturally detour around regions already occupied by critical nets.

%sorted in the ascending order of priority based on large area nets first (reduces future congestion), timing critical (most negative slack) for similar area nets, and then single GCell nets (trivial). 

%\ref{fig:llm_effect}

\subsection{Initial Routing: Steiner / MST + Flexible L-Route}
A preliminary path is assigned to every net for the initial routing pass. Each net's bounding-box area is computed as $A_{n}=(x_{max}^{n}-x_{min}^{n})(y_{max}^{n}-y_{min}^{n})$, and the routing method is selected as follows.

1) Small Nets (Steiner Tree): Nets having area $A_{n}\le2,500$ are routed through RustworkX's Steiner tree algorithm. Steiner tree routing minimizes total wire length to connect all required nodes in the graph, utilizing intermediate GCells as Steiner points to reduce overall cost \cite{chu2008flute}.

2) Large Nets (Prim MST and Flexible L-Route): Nets with area $A_{n}>2,500$ use Prim's minimum spanning tree (MST) on pin terminals, with Manhattan distance $d(p_{1},p_{2})=|x_{1}-x_{2}|+|y_{1}-y_{2}|+40|z_{1}-z_{2}|$. The weight of 40 is the via edge cost in the Manhattan distance to discourage the algorithm from routing through unnecessary layer changes. Each MST edge $(p_{i},p_{j})$ becomes a 2-pin sub-problem solved through flexible L-routing.

3) Flexible dual L-shape routing: The L-shaped paths are evaluated with Horizontal-first and Vertical-first approaches for each 2-pin connection. For each net, the congestion cost is $\Omega(path)=\sum_{(z,y,x)\in path}\max(0,-c_{z,y,x})$, and the L-shape with the lower congestion cost is selected. The complexity for the given algorithm is $O(|A_{n}|)$ cost per net without overhead and latency of searching for the shortest path. The metal layer for routing is assigned by the L-route through maximum remaining capacity among all layers whose segment orientation matches the preferred direction. Specifically, for a horizontal segment the selected layer is $z^* = \arg \max_{z} \bar{c}_{z}$, where $\bar{c}_{z}$ is the mean remaining capacity along the segment's GCell range. This reduces the fixed two-layer assignments and reduces initial overflow by 2.2x on the full ARIANE design $(2,239,467\rightarrow1,029,699)$. Overflow, computed once initial routing completes, is defined as:

\begin{equation}
\Omega=\sum_{z,y,x}\max(0,-c_{z,y,x})
\label{eq:overflow}
\end{equation}

Here $c_{z,y,x}$ is the remaining capacity after routing, which is negative if wire demand is exceeding supply. If $\Omega=0$, the overflow of the circuit is zero and routing is completed; otherwise, it enters the adaptive rip-up and reroute loop.

\subsection{Overflow Analyzer}
We use a SHAP-based approach to decompose the overflow distribution into different parameters before the R\&R iterations.

1) Per-net overflow score: The overflow score for each net routing through GCells is:

\begin{equation}
\psi_{n}=\sum_{(z,y,x)\in\mathcal{G}_{n}}\max(0,-c_{z,y,x})
\label{eq:psi}
\end{equation}

Here $\psi_{n}$ uses the same per-cell overflow measure as $\Omega$ (Eq.~\eqref{eq:overflow}), but is restricted to GCell set $\mathcal{G}_{n}$ occupied by net $n$, such that $\Omega\le\sum_{n}\psi_{n}$ since nets may be sharing congested cells. The nets are ranked by direct contribution to congestion $\psi_{n}$. The top 20\% nets causing overflow are ripped up and rerouted.

2) Layer-wise overflow decomposition: Identifies metal layers having most congestion:

\begin{equation}
\lambda_{z}=\frac{\sum_{y,x}\max(0,-c_{z,y,x})}{\sum_{z,y,x}\max(0,-c_{z,y,x})}.
\end{equation}

The dominant layers $z^{*}=\arg\max_{z}\lambda_{z}$ are passed to the adaptive policy and LLM analyzer.

3) Congested cell fraction: This measures how closely cell utilization predicts overflow and is used to choose from history-driven and present-driven cost inflation.

\begin{equation}
f_{c}=\frac{|\{(z,y,x):c_{z,y,x}<0\}|}{N_{L}\cdot y_{max}\cdot x_{max}}
\end{equation}

$f_{c}$ is congested cell fraction, defined as the ratio of GCells with negative remaining capacity to the total grid size.

4) Utilization-overflow correlation: Cell utilization is defined as $u_{z,y,x}=1-c_{z,y,x}/c_{z,y,x}^{0}$, where $c_{z,y,x}^{0}$ is the original capacity from the cap file. A SHAP-style permutation importance analyzer that decomposes post-routing overflow \cite{lundberg2017unified} is implemented without the external SHAP library; it runs in under 0.1 s on a 4.17 M-cell grid.

\begin{table}[h]
\centering
\caption{Adaptive Policy Rules}
\label{tab:policy}
\begin{tabular}{@{}lll@{}}
\toprule
SHAP signal & Condition & Policy action \\
\midrule
Congested cell fraction & $f_{c}<0.02$ & Increase $p_{f}$ \\
Util-overflow correlation & $r_{uo}>0.5$ & Increase $\alpha$ \\
Layer concentration & $\lambda_{z^{*}}>0.50$ & Decrease via cost \\
\bottomrule
\end{tabular}
\end{table}

The Pearson correlation between cell utilization and the binary overflow is computed over all cells as:

\begin{equation}
r_{uo}=\frac{Cov(u,\max(0,-c))}{\sigma_{u}\cdot\sigma_{\max(0,-c)}}
\end{equation}

A high value of correlation $(r_{uo}>0.5)$ is an indication of cell utilization being a reliable predictor of overflow, forcing the policy to increase the history penalty weight $\alpha$. A low correlation is an indication of congestion being structurally driven (e.g., macro blockages) rather than demand-driven, in which case the history penalty weight is not increased.

\subsection{LLM Log Analyzer}
LLMs are invoked after SHAP analysis is completed. A structured metric prompt is generated and sent to a locally running Ollama instance hosting Llama 3.2 3B. Groq API endpoints hosting Llama 3.3 70B and GPT-OSS 120B were used for the ablation configurations to generate the logs. The prompt contains seven fields: current overflow $\Omega_{L2}$, overflow delta from the initial routing $\Delta\Omega$, congested cell fraction $f_{c}$, utilization-overflow correlation $r_{uo}$, dominant layer overflow percentage $100\lambda_{z^{*}}$, current $\alpha$, and current $p_{f}$. The LLM returns a structured JSON object containing actions and values for $\alpha$, $p_{f}$ and via cost, plus a short natural-language rationale. The values suggested by LLMs are applied after checking validity from hard-coded rules from the knowledge graph, otherwise the SHAP adaptive policy rules described in Table~\ref{tab:policy} are applied.

%%\begin{figure}[t]
%%  \centering
 %% \caption{LLM input/output log.}
  %%\label{fig:llm}
%%\end{figure}

\definecolor{inputblue}{RGB}{45,90,180}
\definecolor{outputgreen}{RGB}{0,120,60}

\begin{figure}[t]
  \centering

\begin{tcolorbox}[
    colback=white,
    colframe=black!20,
    boxrule=0.5pt,
    arc=1mm,
    left=2mm,
    right=2mm,
    top=0.5mm,
    bottom=0.5mm
]

{\scriptsize\ttfamily\setlength{\parskip}{0pt}\linespread{0.9}\selectfont
{\color{inputblue}INPUT: Routing iteration 1 metrics:}

{\color{inputblue}- Overflow (L2): 258,229}
{\color{inputblue}- Overflow delta from init: +0.0\%}
{\color{inputblue}- Congested cells: 1.15\% of grid}
{\color{inputblue}- Util-overflow correlation: 0.237}
{\color{inputblue}- Top layer overflow: 50.1\%}
{\color{inputblue}- Current alpha: 0.4000}
{\color{inputblue}- Current present factor: 2.000}
{\color{inputblue}- Current via cost: 40.0}
{\color{inputblue}- Vias ripped this iteration: 0}

\vspace{0.5mm}

{\color{outputgreen}\{}
{\color{outputgreen}"path\_action": "INCREASE",}
{\color{outputgreen}"pf\_action": "INCREASE",}
{\color{outputgreen}"via\_action": "DECREASE",}
{\color{outputgreen}"alpha\_value": 0.8,}
{\color{outputgreen}"pf\_value": 4.0,}
{\color{outputgreen}"via\_value": 30.0,}
{\color{outputgreen}"reasoning": "High overflow, concentrated congestion, and high top layer overflow indicate need for stronger penalties and via cost reduction to spread routing demand."}
{\color{outputgreen}\}}
}

\end{tcolorbox}
  \caption{Example policy query at iteration~1: structured SHAP metrics sent to the LLM (blue) and the returned JSON parameter update with rationale (green).}
  \label{fig:llm}
\end{figure}

\subsection{Adaptive PathFinder Policy}
The scalar parameters are used to govern the cost model: $\alpha$ (history penalty growth rate), $p_{f}$ (present overflow), and $w_{via}$ (via cost, initial 40).

1) History cost update: The history cost array $h$ is updated for every edge $e$ at the end of iteration $k$:

\begin{equation}
h^{(k+1)}(e)=h^{(k)}(e)+\alpha^{(k)}\cdot \max(0,-c_{e}),
\end{equation}

where $c_{e}=\min(c_{z,y_{1},x_{1}},c_{z,y_{2},x_{2}})$ is the minimum remaining capacity of two cells incident to the edge. The history array is persistent across all R\&R iterations; cells that are repeatedly congested accumulate an increasing penalty.

2) Reroute cost: The Dijkstra algorithm uses combined edge cost during rerouting:

\begin{equation}
cost(e)=W(e)+h^{(k)}(e)+p_{f}^{(k)}\cdot \max(0,-c_{e}),
\label{eq:reroute_cost}
\end{equation}

where $W(e)$ is the weight from Eq.~\eqref{eq:base_weight}. The three additive terms correspond to: the capacity-weighted base cost, accumulated history penalty from past iterations, and the present state overflow penalty $p_{f}$.

3) Automatic alpha growth: $\alpha$ grows by a factor of 1.15 each iteration, $\alpha^{(k+1)}=\alpha^{(k)}\times1.15$, ensuring the history penalty increases even without LLM input.

4) SHAP-driven parameter rules: Table~\ref{tab:policy} showcases three rules for which SHAP can override the default parameter trajectory. $w_{via}$ is decreased when one layer dominates the overflow; this forces the router to use other metal layers.

\subsection{Rip-Up Candidate Selection}
Given the SHAP scores computed by Eq.~\eqref{eq:psi}, the rip-up candidate set for iteration $k$ is:

\begin{equation}
\mathcal{R}^{(k)}=\{n\in\mathcal{N}:\psi_{n}\ge\psi_{\lfloor(1-\rho)|\mathcal{N}|\rfloor}\}
\end{equation}

where $\rho=0.20$ is the rip-up fraction and $\mathcal{N}$ is the set of nets with $\psi>0$. This selects the top 20\% of nets ranked by overflow contribution. The capacity consumed by each ripped net is restored before routing begins. The routing demand of the cell for that given net is incremented for every GCell (z,y,x) that is now occupying the ripped net $n$. This is to ensure that Dijkstra can find the optimum path while considering the capacity of ripped cells. The number of ripped nets decreases monotonically in the full ARIANE design across different iterations.

\subsection{Dijkstra Maze Rerouting}
\label{subsec:maze}
Nets having area between $2,500<A_{n}\le15,000$ use 3D Dijkstra maze routing during the R\&R. A local RustworkX graph is built over the net bounding box with a margin of $\delta=2$ and having edge weights from Eq.~\eqref{eq:reroute_cost}. The shortest path with Dijkstra detours around congested hotspots. Small nets $(A_{n}\le2,500)$ use Steiner trees; large nets $(A_{n}>15,000)$ use MST+FlexL.

\subsection{Knowledge Graph}
The Knowledge Graph is a graph built on RustworkX which keeps a record of routing decisions and outcomes of routing across all R\&R iterations. It keeps a record of three node types:
\begin{itemize}
    \item Pattern: Characteristics of the net at the time of rerouting consisting of net size, dominant layer, and overflow level $\lambda_{z}$.
    \item Strategy: Keeps track of the routing algorithm and active policy parameters $(\alpha^{(k)},p_{f}^{(k)},w_{via}^{(k)})$.
    \item Outcome: the result $\Delta\Omega=\Omega^{(k)}-\Omega^{(k-1)}$ of applying the strategy to the pattern.
\end{itemize}

\section{Experimental Setup}
All experiments were run on a single NVIDIA A40 GPU (48 GB VRAM), 48 GB system RAM, 9 vCPUs, and 40 GB disk. The router and ablation framework were implemented in Python with open-source libraries, and the LLM-enabled runs used the Groq Open Models API for model inference in the policy-analysis loop.

\subsection{Benchmarks}
We evaluate AlphaRoute across benchmarks from the blind dataset of ISPD 2025. The benchmarks are ARIANE (10 layers, $646\times646$ GCells, 105,924 nets), BSG (10 layers, $1384\times1384$ GCells, 768,239 nets), MEMPOOL (10 layers, $1120\times1120$ GCells, 157,744 nets), and NVDLA (10 layers, $386\times386$ GCells, 135,814 nets). All the benchmarks consist of .cap and .net input formats.

\subsection{Ablation Configurations}
Eight cumulative ablations are evaluated. A1 is the Steiner tree baseline. A2 adds flexible L-routing. A3 adds SHAP-driven rip-up and reroute. A4 adds Dijkstra maze rerouting (Section~\ref{subsec:maze}). A5 activates the knowledge graph. A6 adds Graph Neural Network (GNN)-based congestion scoring. A7 and A8 add LLM-guided policy adaptation using Llama 3.3 70B and GPT-OSS 120B, respectively.

\subsection{LLM Ablation}
We compared two model families through the same prompt format and constraints: Llama 3.3 70B Versatile 128k and GPT-OSS 120B 128k, both served via the Groq Open Models API. Across benchmarks, differences in WNS and TNS between models were not statistically meaningful, indicating the policy modification layer is largely model-agnostic since hard constraints and SHAP signals dominate final decisions.

\section{Results}

The routing benchmark used in this work follows the ISPD 2025 Global Routing Contest
specification, which defines the scoring metric $S_\text{orig}$,
the congestion penalty $S_\text{overflow}$ and the runtime penalty score
$S_\text{scaled}$. AlphaRoute achieves $S_\text{orig} = 0.0538$ and $S_\text{scaled} = 0.0589$ in ARIANE, representing
a $33.1\times$ reduction in the original score and a $29.5\times$ reduction in the scaled
score relative to the current SOTA values of $S_\text{orig} = 1.780$ and
$S_\text{scaled} = 1.74$ (ARIANE), demonstrating that the overflow reduction delivered by
AlphaRoute decisively outweighs its runtime overhead in the final metric.
Our wall-clock runtime is approximately $136\times$ above the contest median
($t_\text{med} = 19\,\text{s}$), which can be attributed to two compounding factors: (1) our implementation is written entirely in Python rather than the optimized C++ employed by top contest entries, and (2) our experiments used a more constrained
cloud instance than the contest reference hardware (200\,GB RAM, 8 CPU cores, NVIDIA A100 (PG506-230) with 40\,GB GPU memory). We acknowledge this runtime overhead
as a limitation; future work will address it through a C++/Rust re-implementation or GPU-accelerated path search.

\subsection{Overflow Analysis}

The overflow is presented across eight ablations and four benchmark circuits. The overflow reduction is consistent as each component is progressively added.

The overflow score for baseline A1 is the highest overflow across all of the benchmarks with 1{,}517{,}789 for ARIANE, 7{,}600{,}283 for
BSG, 124{,}450 for MEMPOOL and 19{,}297{,}016 for NVDLA. L-routing (A2) reduces it drastically by $82.9\%$ for ARIANE, $79.1\%$ for MEMPOOL, $43.6\%$ for BSG and $43.3\%$ for NVDLA. This conveys that L-routing very efficiently routes large nets.

In A3, SHAP-based rip-up and reroutes are added for nets causing overflow. This reduces overflow for ARIANE to $\sim$241{,}811 and MEMPOOL to $\sim$19{,}438. However, BSG and NVDLA are comparable with A2, indicating that SHAP-driven rip-up and rerouting is more effective on circuits having concentrated congestion hotspots.

The addition of Dijkstra maze routing in A4 produces the second major improvement, as the reduction in ARIANE is $6.3\%$, MEMPOOL is $14.2\%$ and NVDLA is $35\%$. BSG remains unchanged compared to A3, which suggests that structural congestion may be resistant to maze routing alone.

The overflow reduction from A5 through A8 remains marginal. The similar overflow between A4 and A5 proves that the knowledge graph is only used for recording previous iterations' policy parameters.
Ablation of GNN achieves a small improvement on MEMPOOL to $\sim$1{,}726.

\begin{figure}[H]
    \centering
    \includegraphics[width=0.9\linewidth,trim=0cm 1cm 0cm 0cm,clip]{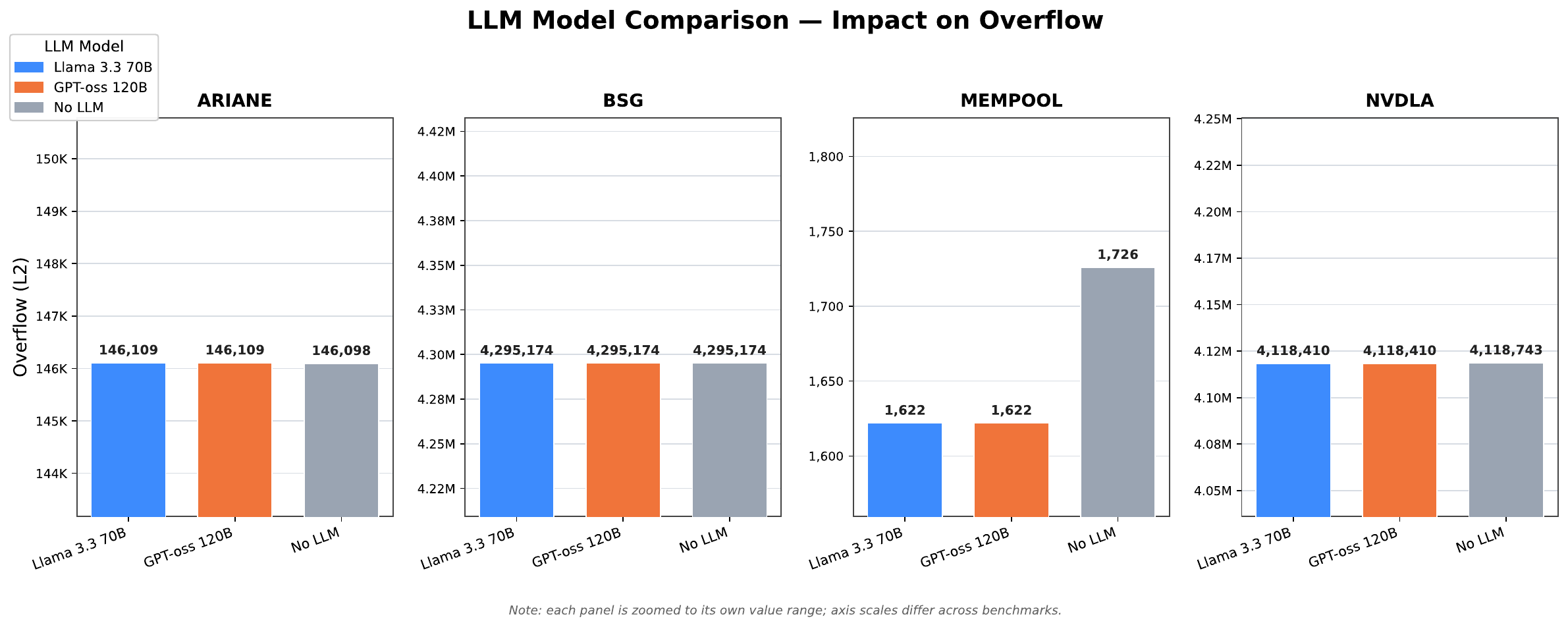}
    
    \caption{Overflow with and without LLM-guided policy adaptation (A6 vs.\ A7/A8); LLMs slightly over-penalize designs already near optimal, contributing interpretability rather than further reduction.}
    \label{fig:llm_effect}
\end{figure}

Further, LLM-guided policy adaptation yields identical results to A4 and A5 and proves that overflow reduction is saturated and LLMs contribute more to interpretability than to direct overflow reductions. The slight increase in overflow for A7 and A8 is due to LLM-suggested parameter adjustments that slightly over-penalize congested GCells on a design that has already reached near-optimal routing under GNN ordering, as shown in the plot of the impact of overflow due to LLMs in Fig.~\ref{fig:llm_effect}.

Overall, the results demonstrate that flexible L-routing~(A2), SHAP-guided
rip-up~(A3), and Dijkstra maze routing~(A4) account for most overflow reductions, while GNNs and LLMs provide interpretability.
\begin{figure}[H]
    \centering
    \includegraphics[width=0.9\linewidth,trim=0cm 0cm 12cm 0cm,clip]{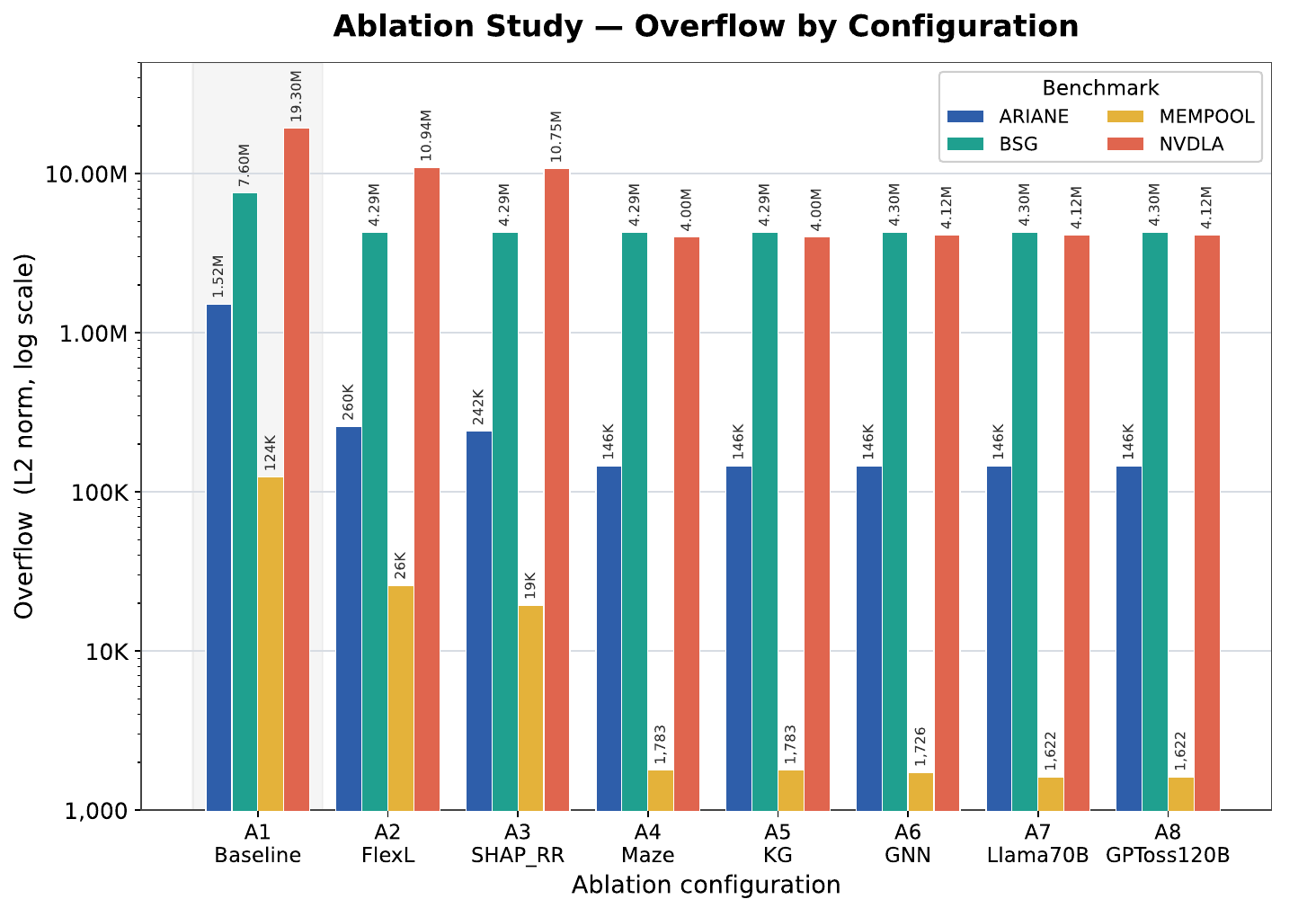}
    \caption{Absolute overflow across cumulative ablations A1--A8 for all four benchmarks; most reduction is delivered by L-routing (A2), SHAP rip-up (A3), and maze routing (A4).}
    \label{fig:abs_overflow}
\end{figure}
\subsection{R\&R Convergence Analysis}
\begin{figure}[t]
  \centering
  \begin{subfigure}[b]{0.48\linewidth}
    \includegraphics[width=\linewidth]{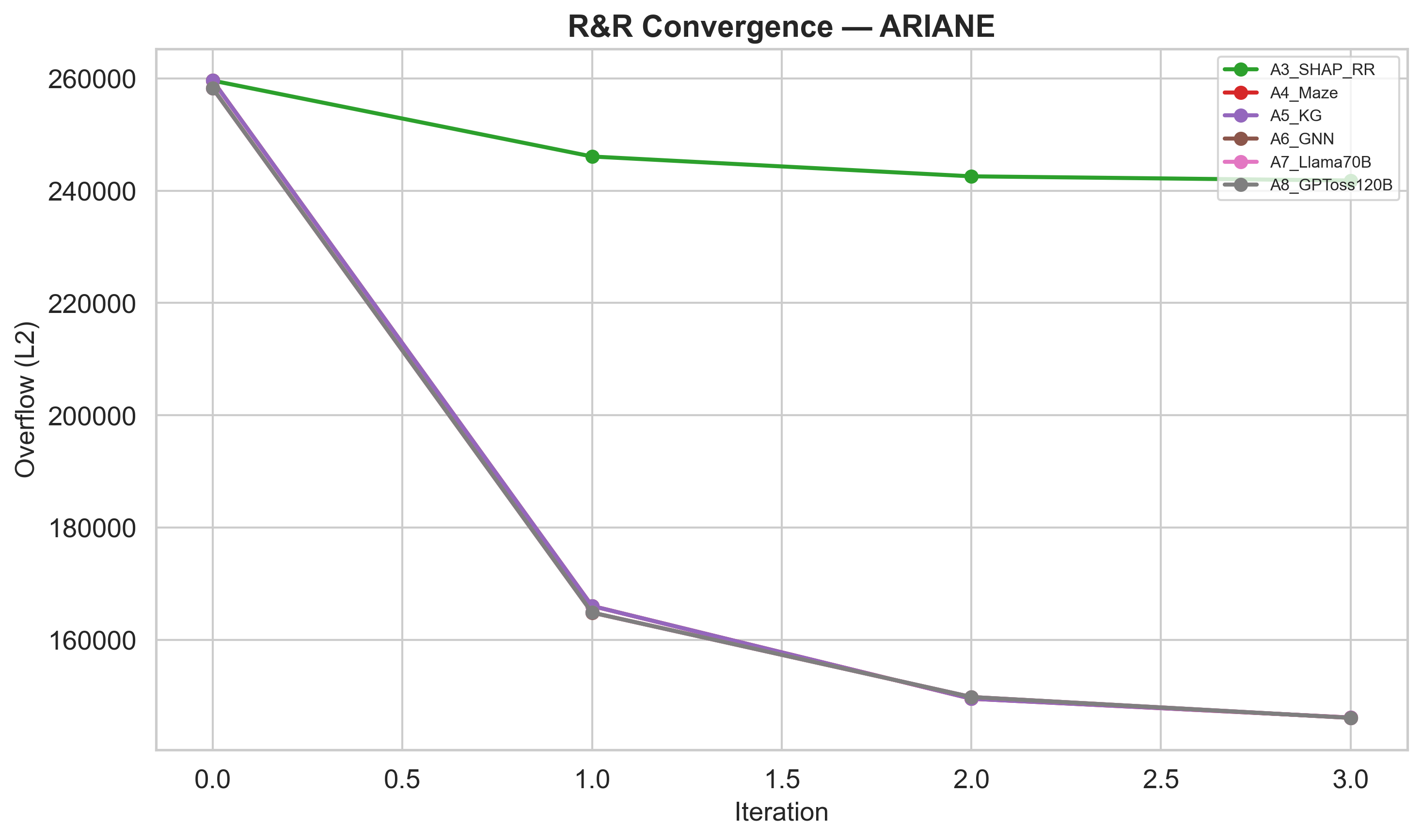}
    \caption{ARIANE}
    \label{fig:conv_ariane}
  \end{subfigure}
  \hfill
  \begin{subfigure}[b]{0.48\linewidth}
    \includegraphics[width=\linewidth]{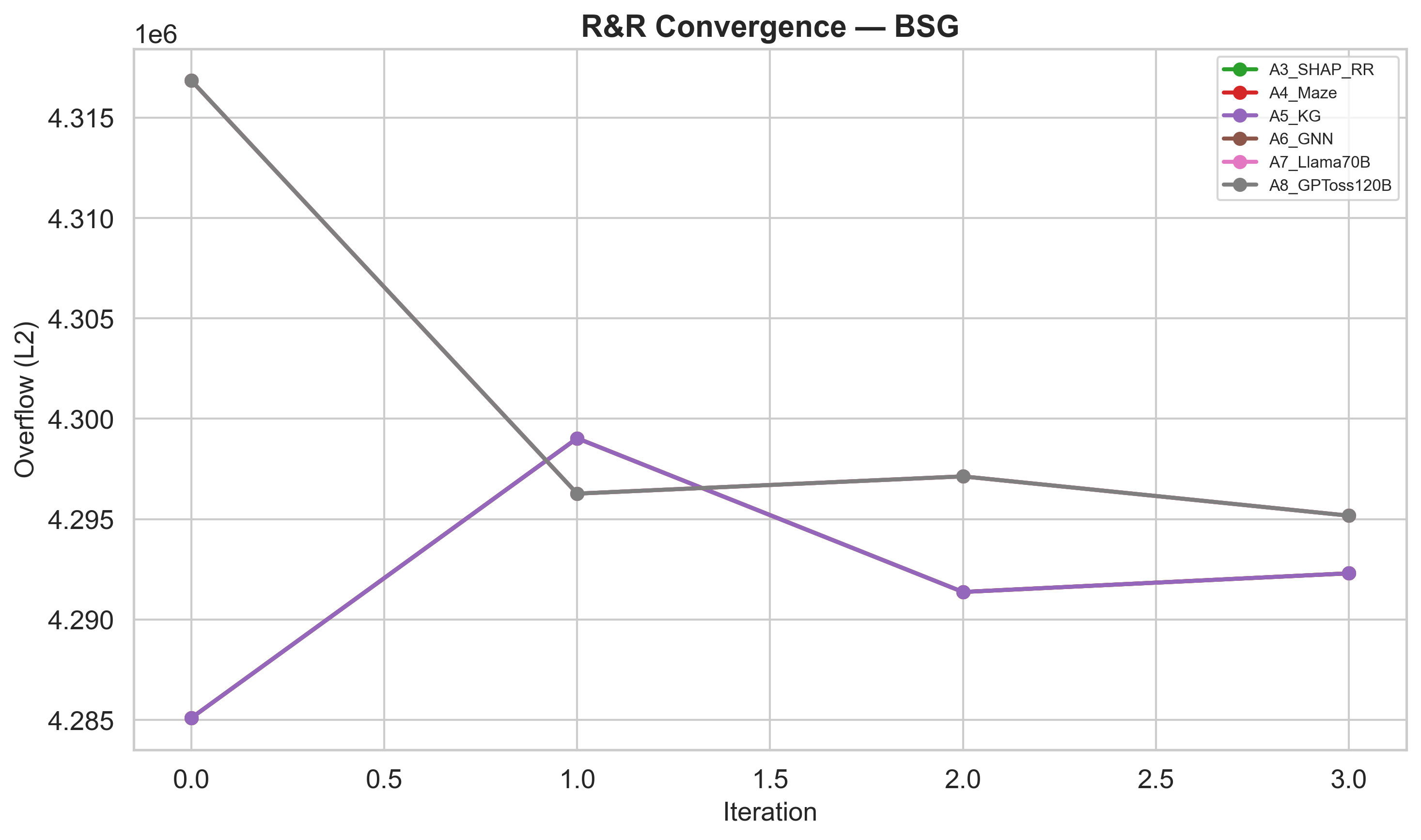}
    \caption{BSG}
    \label{fig:conv_bsg}
  \end{subfigure}

  \medskip

  \begin{subfigure}[b]{0.48\linewidth}
    \includegraphics[width=\linewidth]{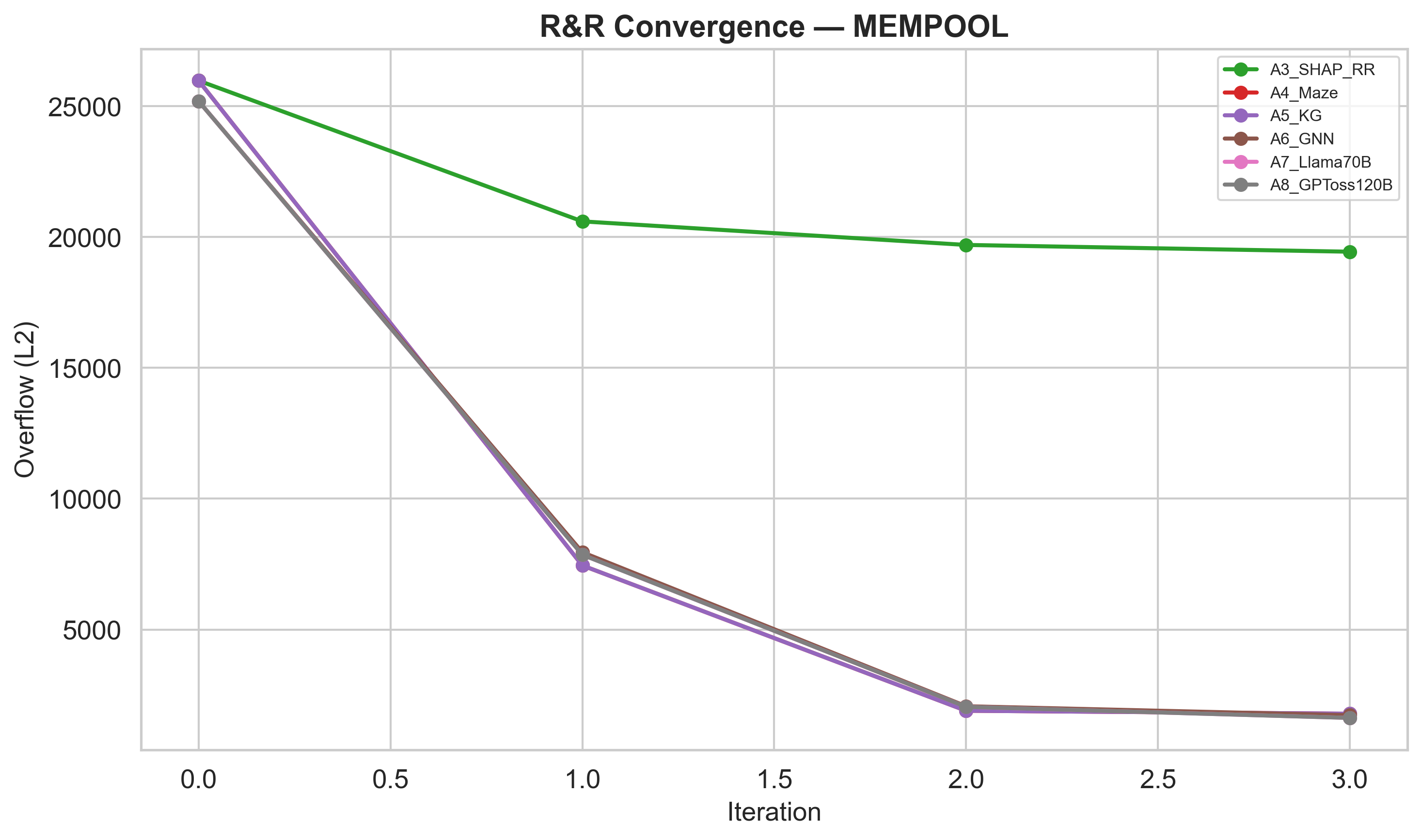}
    \caption{MEMPOOL}
    \label{fig:conv_mempool}
  \end{subfigure}
  \hfill
  \begin{subfigure}[b]{0.48\linewidth}
    \includegraphics[width=0.8\linewidth]{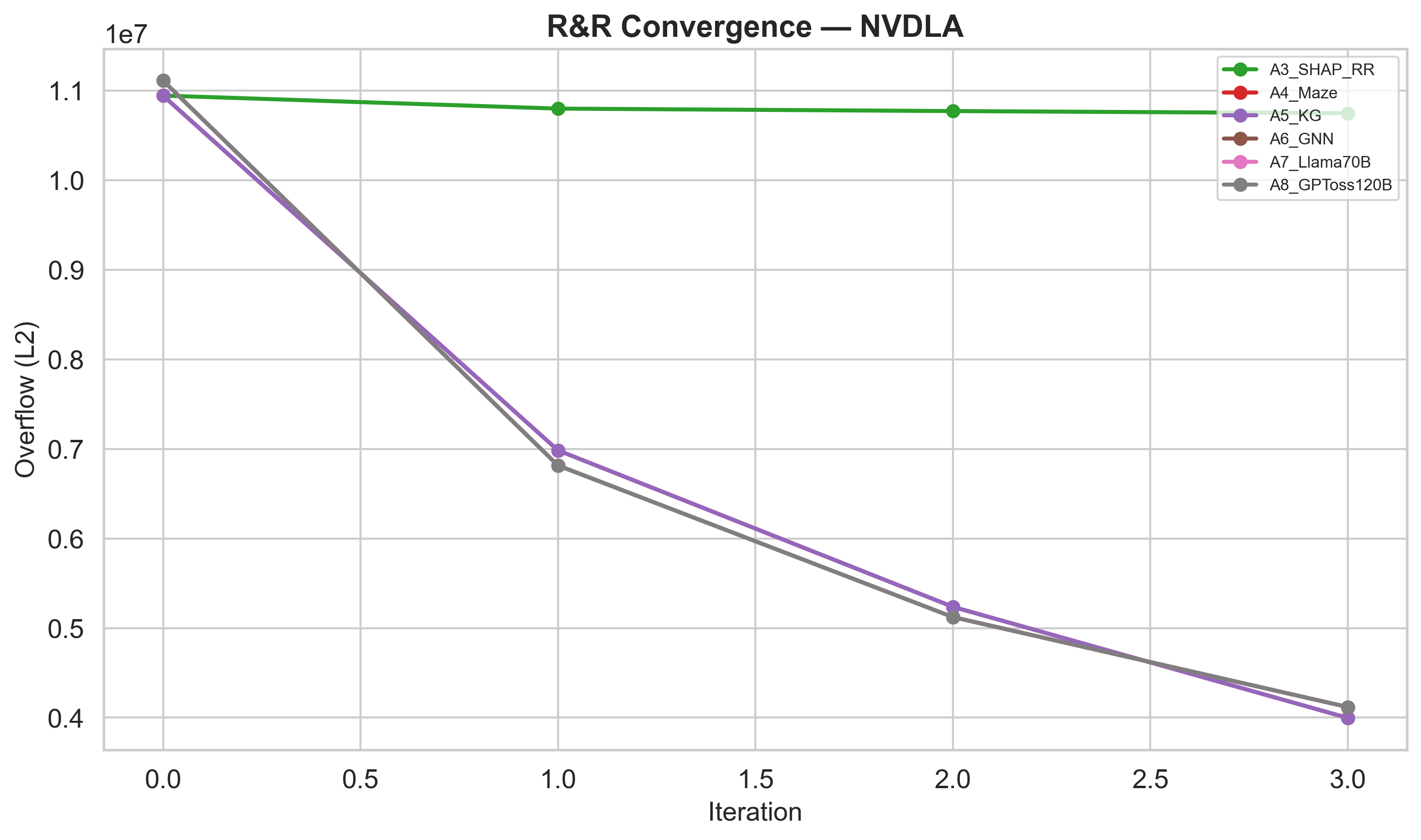}
    \caption{NVDLA}
    \label{fig:conv_nvdla}
  \end{subfigure}
  \caption{R\&R convergence curves (overflow L2 vs.\ iteration)
           for configurations A3--A8 across all four benchmarks.}
  \label{fig:convergence}
\end{figure}

Fig.~\ref{fig:convergence} shows overflow convergence over three R\&R iterations for A3--A8. On ARIANE, A4--A8 converge steeply from $\sim$260{,}000 at iteration~0 to $\sim$148{,}000 by iteration~3; MEMPOOL follows the same trend, dropping from $\sim$25{,}000 to below $\sim$2{,}000 (a $92\%$ reduction) by iteration~2, and NVDLA falls from $\sim$1.1$\times$10$^7$ to $\sim$4.0$\times$10$^6$. In every case A3 plateaus early, confirming that Dijkstra maze routing is needed to escape local congestion minima on large or structurally congested designs.

BSG illustrates the LLM's role: A8 converges sub-linearly as the LLM shifts from overflow reduction to logged reasoning once routing nears its optimum, so LLM guidance helps most when residual congestion is significant and alternative paths exist. The near-identical A4--A7 trajectories on ARIANE and MEMPOOL confirm that the KG, GNN, and LLM layers do not alter convergence dynamics.

\section{Conclusion}
AlphaRoute demonstrates that SHAP-driven overflow decomposition, adaptive PathFinder tuning, Dijkstra maze rerouting, and LLM-guided policy adaptation can be composed into a single closed-loop system that reduces congestion by up to two orders of magnitude on ISPD 2025 benchmarks, achieving an original score over 33x lower than the published state of the art on ARIANE despite a pure Python implementation. Ablations confirm that flexible L-routing, SHAP-targeted rip-up, and maze rerouting account for nearly all overflow reduction, while the knowledge graph and LLM layers provide interpretability without degrading routing quality, and the policy loop remains model-agnostic across LLM families. Future work includes porting critical loops to compiled code or GPU kernels to close the runtime gap, distilling a compact routing-specific language model from accumulated knowledge-graph logs, and extending the SHAP analyzer to capture net-to-net coupling for structurally congested designs where per-net marginal scores are least effective.

\bibliographystyle{IEEEtran}
\bibliography{refs}

\end{document}